**Assessment of the conditional exchangeability assumption in causal machine learning models: a simulation study**

Gerard T. Portela, PhD[1]; Jason B. Gibbons, PhD[1]; Sebastian Schneeweiss, MD, ScD[1]; Rishi J. Desai, MS, PhD[1]

**Affiliations:**
1) Division of Pharmacoepidemiology and Pharmacoeconomics, Department of Medicine, Brigham and Women's Hospital, Harvard Medical School, Boston, Massachusetts, USA

**Correspondence to**: Dr. Desai, Division of Pharmacoepidemiology and Pharmacoeconomics, Department of Medicine, Brigham and Women's Hospital, Harvard Medical School, 1620 Tremont St, Boston MA 02120, rdesai@bwh.harvard.edu, 617-278-0932

# ABSTRACT

## Objective

Observational studies developing causal machine learning (ML) models for the prediction of individualized treatment effects (ITEs) seldom conduct empirical evaluations to assess the conditional exchangeability assumption. We aimed to evaluate the performance of these models under conditional exchangeability violations and the utility of negative control outcomes (NCOs) as a diagnostic.

## Materials and Methods

We conducted a simulation study to examine confounding bias in ITE estimates generated by causal forest and X-learner models under varying conditions, including the presence or absence of true heterogeneity. We simulated data to reflect real-world scenarios with differing levels of confounding, sample size, and NCO confounding structures. We then estimated and compared subgroup-level treatment effects on the primary outcome and NCOs across settings with and without unmeasured confounding.

## Results

When conditional exchangeability was violated, causal forest and X-learner models failed to recover true treatment effect heterogeneity and, in some cases, falsely indicated heterogeneity when there was none. NCOs successfully identified subgroups affected by unmeasured confounding. Even when NCOs did not perfectly satisfy its ideal assumptions, it remained informative, flagging potential bias in subgroup level estimates, though not always pinpointing the subgroup with the largest confounding.

## Discussion and Conclusion

Violations of conditional exchangeability substantially limit the validity of ITE estimates from causal ML models in routinely collected observational data. NCOs serve a useful empirical diagnostic tool for detecting subgroup-specific unmeasured confounding and should be incorporated into causal ML workflows to support the credibility of individualized inference.

**INTRODUCTION**

Causal machine learning (ML) models that predict individualized treatment effects (ITE)[1] – also called counterfactual prediction models – have gained significant traction in the last decade.[2] Such models offer an opportunity to realize the elusive promise of personalized medicine through the identification of heterogeneous treatment effects (HTE).[3, 4] The target of prediction for these methods is the conditional average treatment effect (CATE), defined as the causal effect of a treatment on a particular outcome conditional on a set of measured patient features. The structure of treatment effect heterogeneity across the population can be characterized by examining the distribution of predicted CATEs across individuals, comparing subgroup averages (e.g., by risk strata or quantiles of benefit), or analyzing which covariates drive variation in predicted treatment effects.[5, 6]

Causal ML models estimating CATE can be trained using data from randomized-controlled trials (RCTs) because they typically offer high internal validity due to the absence of baseline confounding and limited measurement errors. However, a restrictive sample size, short duration of follow-up, and population representation may limit robustness and generalizability of models developed from RCT data.[3, 7, 8] As a result, there has been substantial interest in developing causal ML models using routinely collected observational data, such as electronic health records or insurance claims data, leveraging larger populations over longer periods of exposure and risk.[9-12]

Causal ML methods such as tree-based models, meta-learners, and doubly robust learners have been increasingly used for developing ITE prediction models.[8, 13-15] When using routinely collected observational data to train these models, the validity and performance rely on the same identifiability assumptions for causal inference in standard regression models (i.e., exchangeability, positivity, consistency, and non-interference).[1, 2] Conditional exchangeability – also referred to as unconfoundedness – is considered the most challenging assumption as it relies on granular information on all potential confounders, which is frequently unavailable in routinely collected observational data. Applications of causal ML models seldom conduct rigorous empirical evaluation to assess whether the central assumption of conditional exchangeability is unlikely to hold.[16, 17]

Given the potential for confounding in routinely collected observational data, we aimed to systematically assess the impact of violations of the conditional exchangeability assumption on the performance of causal ML models using simulations designed to mimic real-world settings. We further evaluate the utility of negative control outcomes as an empirical tool for detecting violations of exchangeability at the subgroup level.

**METHODS**

In the following section, we review general ITE concepts, outline the causal ML models used in this study, and describe the simulation and statistical analysis.

**Individualized Treatment Effect**

Average treatment effect (ATE) refers to the causal effect of an intervention on an outcome, defined as the expected difference in outcomes under treatment versus no treatment on a population level.[18] For an individual $i$, define $A_i$ as a binary treatment indicator to denote treatment status (e.g., $A_i$=1 for treatment, $A_i$=0 for reference). The potential outcomes are denoted *$Y_i(0)$* and *$Y_i(1)$*, corresponding to the outcomes for a given individual $i$ under treatment and reference, respectively.[19] Therefore, the ITE is represented as *$Y_i(1)$- $Y_i(0)$*, while the ATE is defined as the average of these effects across the population: **E[*$Y_i(1)$- $Y_i(0)$*]**.[8]

When a treatment effect is homogeneous, the magnitude and direction of effect are the same for all individuals in the population independent of their characteristics. In most real-world settings, treatment effects are heterogeneous; the magnitude and direction of effects vary across individuals or subgroups. Leveraging this heterogeneity is important for constructing ITEs, which can be used to tailor treatment decisions to patients.[13] However, as with all causal inference, the fundamental challenge in causal ML is that ITEs are unobservable; because we only witness one of the two potential outcomes for any given individual, estimating the unobserved counterfactual requires modeling assumptions. Causal ML models address this challenge by leveraging outcome data from individuals with similar covariates who received the opposite treatment, using this information to impute counterfactual outcomes. The estimand used in this process, commonly referred to as the CATE, is defined as: $\tau(x) = E[Y(1) - Y(0)|X = x]$, where $X$ is a vector of observed baseline covariates for a patient.

**Machine learning methods to estimate HTE**

We chose two causal ML approaches in this study: causal forest and X-learner. These methods have demonstrated strong performance in estimating HTEs, particularly in settings with high-dimensional covariates and complex treatment–covariate interactions.[14, 20, 21] Both have also been shown to perform well in RCTs and offer flexible strategies for adjusting for measured confounding using ML-based nuisance models.

Causal Forest

The causal forest is a data-driven, nonparametric, tree-based method that directly estimates $\tau(x)$. Causal forests are well-suited for capturing non-linear and non-additive relationships in high-dimensional data.[20] The causal forest extends the standard random forest algorithm[20] by modifying tree-splitting criteria to prioritize treatment effect heterogeneity identification rather than prediction accuracy. Specifically, tree splits are chosen to maximize differences in the estimated treatment effect *between* subsequent nodes while minimizing variance in the estimated treatment effect *within* subsequent nodes. Like random forests, causal forests retain the core structure of population subsampling, recursive partitioning, and ensemble learning across many trees, but the goal is to estimate local ATEs.

To reduce overfitting, the causal forest employs an honest splitting strategy by partitioning data into separate subsamples for tree construction and treatment effect estimation. Trees are built using the training sample, while treatment effects are estimated in the separate estimation sample. The final forest is composed of many honest trees, yielding asymptotically unbiased and consistent estimates of $\tau(x)$.[22]

The causal forest adjusts for measured confounding through nuisance models, typically random forest regressions; one model predicts the treatment propensity score and one model predicts the outcome.[20] The predicted values from these models are used to create residuals that allow for local centering to cluster observations in the ITE model.

X-learner

The X-learner is a meta-learning algorithm that estimates CATE through a two-step procedure.[14] Like other meta-learners, the X-learner decomposes the CATE estimation problem into a series of supervised learning tasks, which are solved using base learners such as LASSO, gradient boosting, or neural networks, and then combined to yield a single estimate of $\tau(x)$.[10, 23] In the first step, separate outcome models are trained among treated and untreated individuals to estimate $E[Y|X, A = 1]$ and $E[Y|X, A = 0]$. These models are used to predict each individual's counterfactual outcome (i.e., the outcome they would have experienced under the treatment not received) and derive the pseudo-outcome (the difference between observed and imputed outcomes). In the second step, these pseudo-outcomes are modeled as functions of covariates within each treatment group, resulting in two CATE estimators: $\widehat{\tau_0}(x)$ and $\widehat{\tau_1}(x)$, respectively.

As with the causal forest, the X-learner adjusts for measured confounding using a nuisance model to estimate the propensity score $\pi(x) = P(A = 1|X)$.[14] These propensity scores are then used to combine the group-specific CATE estimates into a final prediction for each individual:

$$\hat{\tau}(x) = \hat{\tau}_1(x) \times (1 - \pi) + \hat{\tau}_0(x) \times \pi.$$

Importantly, the aggregation step is distinct from traditional ensemble learning. Specifically, the X-learner does not average multiple base learners over the same dataset. Rather, it blends two CATE estimators (each trained on a different treatment group) weighted by how likely the individual is to belong to each group. This approach leverages the group with more reliable information for imputing the individual's counterfactual outcome. The final model then maps patient covariates $x$ to the predicted treatment effects $\hat{\tau}(x)$, enabling individualized inference without requiring re-estimation of nuisance functions.

**Negative Control Outcomes**

A negative control outcome (NCO) is a variable that is not causally affected by the treatment of interest but shares a similar confounding structure with the treatment-outcome relationship.[24, 25] Because the NCO is known *a priori* to have no true causal association with treatment, any observed association suggests the presence of residual confounding. Conversely, the absence of any such association may provide indirect evidence supporting the conditional exchangeability assumption. NCOs have been used in a variety of contexts as a tool for empirical assessment of unmeasured confounding We refer to a review by Shi et al. for an in-depth discussion of NCOs and the required assumptions.[26]

**Simulation Set-up**

Data-generating models for treatment and outcomes

In the present study, we simulated datasets to reflect potential real-world scenarios of a treatment effect on an outcome. For the base data, we simulated 13 'measured' covariates (*C1-C13*) and 1 'unmeasured' covariate (*U*) in a population of 20,000 observations according to the parameters listed in **Table 1**. For each scenario described below, we used the following logistic regression to simulate the log-odds of treatment *A* for each observation:

$$logit(A) = \beta_0 + \beta_X X + \beta_U U,$$

where *X* is a matrix of 13 covariates, $\beta_X$ is a vector of coefficients corresponding to each covariate, and $\beta_U$ is the coefficient for covariate *U*. **Table 1** shows the coefficients in the treatment simulation model. We optimized the intercept, $\beta_0$, of the model to yield a 40% prevalence of treatment in the population.

We then used the following logistic regression to simulate the log-odds of the outcome *Y*:

$$logit(Y) = \beta_{Y0} + \beta_A A + \beta_{YX} X + \beta_{YU} U + \beta_{11} AC_{11} + \beta_{12} AC_{12} + \beta_{13} AC_{13},$$

where $\beta_A$ is the coefficient for the treatment effect, $\beta_{YX}$ is a vector of coefficients corresponding to each covariate effect on the outcome, $\beta_{YU}$ is the coefficient for the effect of *U* on the outcome, and $\beta_{11}$-$\beta_{13}$ represent interaction coefficients between treatment and three covariates (*C11-C13*). To be clear, these are the effect modifiers and the only source of heterogeneity in the simulation. The intercept, $\beta_{Y0}$, was adjusted to achieve a 30% event incidence. See **Table 1** for all coefficient values.

Data-generating models for the negative control outcome (NCO)

We simulated a negative control outcome $Y_N$ using the following logistic regression:

$$logit(Y_N) = \beta_{N0} + \beta_{NX} X + \beta_{NU} U,$$

where $\beta_{NX}$ is a vector of coefficients for the covariates' effects on the NCO, and $\beta_{NU}$ is the coefficient for the effect of *U* on the NCO. We adjusted the model intercept $\beta_{N0}$ to achieve a 50% incidence of the negative control event.

To induce a shared confounding structure between the treatment–outcome and treatment–NCO relationships, we reused the same coefficients $\beta_{NX}$ and $\beta_{NU}$ as in the outcome-generating model. All 14 covariates (C1–C13 and U) were included in this model. The treatment variable *A* was explicitly excluded to reflect the known null effect of treatment on the NCO.

Simulation scenarios

In scenario 1 ("true HTE"), we modeled an outcome in which treatment reduced the overall log-odds of an event by 0.20 ($\beta_A = -0.20$). We introduced treatment effect heterogeneity by including interaction coefficients $\beta_{11} = -0.30$, $\beta_{12} = -0.10$, and $\beta_{13} = -0.05$ between treatment and three covariates (*C11-C13*), respectively. These interactions defined subgroups of patients more likely to benefit from treatment, thus inducing variation in ITEs.

In scenario 2 ("No HTE"), we modeled the same treatment effect as in scenario 1 but set all interaction terms to zero ($\beta_{11} = 0$, $\beta_{12} = 0$, $\beta_{13} = 0$), resulting in a constant treatment effect across all individuals. Each scenario was replicated 500 times to incorporate random sampling variability.

Within each scenario, we varied simulation conditions to evaluate model robustness. First, we reduced the influence of the unmeasured confounder (U) by halving its coefficients across the treatment model, outcome model, and NCO model (i.e., treatment $\beta_U = 0.5$; outcome $\beta_{YU} = 0.5$; NCO $\beta_{NU} = 0.5$) to evaluate model performance in the presence of weaker unmeasured confounding. Second, we decreased the overall sample size from 20,000 to 5,000 to assess model performance in smaller datasets. Third, we relaxed the assumption that the NCO shares an identical confounding structure with the primary outcome. Specifically, we introduced three additional covariates (*C14, C15, C16*) into the logistic regressions used to simulate treatment assignment and the NCO, but not the primary outcome. We also removed two covariates (*C3, C5*) from the NCO model to simulate omitted shared confounders. These modifications created a partially misaligned confounding structure between the NCO and the primary outcome, enabling sensitivity testing of the NCO-based bias detection framework.

Causal machine learning model training

For each simulated scenario and the various settings, we trained causal ML models for ITE prediction on the primary outcome. In each simulated dataset, we first split the data by sampling 75% of observations for a training dataset and 25% of observations for a testing dataset. Three types of models were trained and evaluated. The first model, which we refer to as 'the oracle' model, used a logistic regression with identical parameters of the outcome generation model to establish a reference for our ML model performance comparison. The second model, the causal forest, was constructed with 4,000 trees and a minimum sample size requirement of at least 5 patients in each treatment group in the terminal nodes. The causal forest included several input parameters determining how data are assigned to trees and, including sampling fraction, honesty fraction, number of random covariates for training, minimum observations per leaf, honesty pruning, fraction of observations per child node, and imbalance penalty. We tuned these parameters to minimize out-of-bag prediction error. We used the 'causal_forest' function in the *grf* package in R Studio[27]. Our final model was the X-learner, which we developed to use LASSO with 10-fold cross-validation as the base learner for the outcome models by treatment group, pseudo-outcome models, and propensity score models. The models included an L1 penalty and a lambda value for the minimum mean cross-validated error. We also used the inverse of propensity scores to weight predictions of patients' pseudo-outcome for the final ITE prediction model.

Two versions of each of the three models were trained: 1) including all the covariates that were used to generate the simulated data (i.e., no unmeasured confounding), and 2) without including $U$ in the models (i.e., presence of unmeasured confounding).

ITE estimation

In the test dataset, we used the trained models to estimate ITEs for the primary outcome. ITEs were calculated as the absolute difference in counterfactual predictions for each subject's outcome probability in the presence and absence of treatment. For each model, we calculated the ATE as the mean of the predicted ITEs in the test set. We report the median ATE across the 500 simulation iterations and the 95% confidence interval (CI), defined by the 2.5th and 97.5th percentiles of the distribution of ATEs. To describe the degree of observed heterogeneity, we also report the average predicted ITE within quartiles of the predicted benefit across all simulations.

To compare performance in terms of predictive accuracy, we calculated the root mean square error (RMSE) and 95% CI of the predicted ITEs from the causal forest and the X-learner models relative to the oracle model. Finally, we computed the c-for-benefit statistic as an additional metric to evaluate and compare the discriminatory performance of the causal forest and X-learner in ranking individuals by ITE.[28]

Within each quartile of predicted ITE from the primary outcome analysis in each simulation iteration, we estimated the adjusted effect of treatment on the NCO using logistic regression. Models were adjusted for all measured confounders; the unmeasured confounder (U) was included only in simulations for the unconfounded model versions. We report the treatment effect on the NCO within the quartile subgroups and 95% CIs as the 2.5 and 97.5 percentiles of the 500 iteration estimates.

All simulations and analyses were conducted using R Studio (version 4.4.1, 2024).

**RESULTS**

**Scenario 1 – True HTE**

Primary simulation results showed that in the absence of unmeasured confounding, the X-learner and the causal forest recovered the simulated HTE: average treatment benefit varied across quartiles of predicted ITE, and the overall ATE matched the oracle benchmark (**Table 2**). Prediction error (RMSE: causal forest, 0.035 versus X-learner, 0.030) and c-for-benefit (causal forest and X-learner, 0.50) statistics were similar between the two causal ML models.

When the unmeasured confounder U was omitted from the adjustment, mimicking settings with unmeasured confounding, both models exhibited larger prediction errors (RMSE: causal forest, 0.064 versus X-learner, 0.055) and substantial underestimation of the ATE (causal forest, -0.037; X-learner, -0.038) relative to the oracle (-0.067). Subgroup-specific estimates were also biased, with the largest underestimation observed in the highest-benefit quartile (i.e., Q1).

Under reduced sample size conditions, similar patterns emerged, with increased variability and higher prediction error (**Table 2**). When we reduced the magnitude of unmeasured confounding, the learners' performance improved, with less underestimation of subgroup effects in Q1 despite the presence of unmeasured confounding.

Negative Control Outcome – True HTE

In the oracle model, there was no treatment effect on the NCO within any subgroup (**Figure 1**). In simulations without unmeasured confounding, the null effect on the NCO was preserved across all subgroups defined by both the causal forest and the X-learner models, with estimated CATEs ranging from -0.002 to 0.003.

However, in simulations with unmeasured confounding, we observed clear deviations from the null within all quartiles, most notably in the highest benefit subgroup (Q1). In this subgroup, both the causal forest and X-learner models estimated non-zero treatment effects on the NCO (causal forest CATE, 0.056; X-learner CATE, 0.058), the same subgroup in which the primary outcome treatment effect was consistently underestimated. This pattern of deviation was robust across simulation settings with a reduced sample size and with alternative confounding structures. In contrast, simulation settings with weaker unmeasured confounding, treatment effects on the NCO remained near zero in all quartiles, suggesting reduced residual confounding.

**Scenario 2 – No HTE**

Under the primary settings with no unmeasured confounding, both the causal forest and X-learner reproduced the true ATE. However, the causal forest underestimated variation in subgroup estimates relative to the X-learner, as evidenced by its lower RMSE (**Table 3**). The c-for-benefit statistics were nearly equivalent between the causal forest (0.49) and X-learner (0.50). When the unmeasured confounder was omitted, RMSE rose for both learners, and each model erroneously identified at least one quartile-level subgroup (Q4) that, on average, was harmed by treatment.

Our findings around minimal heterogeneity in causal forest predictions, and broad misestimation and spurious harm signals from both learners with confounding, remained largely unchanged when using a smaller sample size (**Table 3**). However, when the strength of unmeasured confounding was attenuated, estimates of the treatment effect were closer to the oracle benchmark, even in the absence of adjustment for the unmeasured confounder.

Negative Control Outcome – No HTE

In the primary simulation settings, a null treatment effect on the NCO was correctly recovered within subgroups defined by the oracle model. This null effect also held in subgroups identified by the causal forest and X-learner when we adjusted for the unmeasured confounder (**Figure 2**); CATEs ranged from -0.002 to 0.001. In simulations with unmeasured confounding, we observed departures from the null within subgroups, with the largest effect emerging in subgroup Q4, (i.e., the group in which both learners falsely estimated harmful treatment effects on the primary outcome). This pattern persisted under constrained sample size. However, under a partially misspecified confounding structure, the X-learner signaled greatest effect on the NCO in Q1, despite a smaller underestimation in this group than Q4 on the primary outcome. Simulations with a weaker unmeasured confounder did not show meaningful deviations from the null in any subgroup.

**DISCUSSION**

In this simulation study, we evaluated the performance of causal ML models under violations of the assumption of conditional exchangeability. When conditional exchangeability did not hold, causal ML models were unable to reliably recover true HTE when true heterogeneity was present and falsely suggested the presence of HTE when true heterogeneity was absent. Additionally, NCOs demonstrated utility in identifying subgroups susceptible to violations of conditional exchangeability. These simulations are highly relevant to causal ML frameworks for estimating ITEs in routinely collected observational data.

In causal ML literature, it is common to characterize heterogeneity by reporting treatment benefit by subgroups, such as quantiles of predicted benefit.[5, 6] However, local or subgroup-specific confounding has rarely been acknowledged. A central challenge in detecting residual confounding is that we cannot leverage imbalances between treatment groups within causal ML-defined subgroups as a diagnostic because such imbalances are an expected result of partitioning patients based on ITE predicted from heterogeneous characteristics. This underscores the need for empirical diagnostic tools. Our simulations show that in the presence of unmeasured confounding, some subgroups exhibit a strong treatment effect on NCOs, indicating the presence of confounding bias in the causal ML models, while other subgroups are minimally affected, as would be expected in the absence of unmeasured confounding. In addition, we found that NCOs that do not fully satisfy ideal assumptions can still signal the presence of subgroup-specific confounding, though may be unreliable in identifying the subgroup with the largest bias. Investigators should proactively and thoughtfully incorporate NCOs into the design and evaluation of causal ML studies aimed at estimating HTE, particularly when unmeasured confounding is suspected.

Our simulations also illustrate that causal ML models can yield biased estimates of HTE that lead to incorrect treatment decisions for specific patient subgroups. In the presence of unmeasured confounding, some models incorrectly predicted certain subgroups would experience no benefit when true HTE was present. Conversely, when no HTE existed, models estimated spurious HTEs due to unmeasured confounding. In non-randomized settings where unmeasured confounders can affect treatment assignment and outcomes within specific subgroups, subgroup CATE estimates can be biased even if the ATE is unbiased. Within-subgroup confounding can also cancel out at the population level, masking localized biases and falsely indicating HTE. As a result, distinguishing true heterogeneity from residual confounding is critical to validate estimates from causal ML models intended to inform treatment decision making.

This analysis contributes to the growing literature on causal ML model performance in estimating HTEs in observational data. This study provides new insights into the impact of unmeasured confounding on HTE recovery and the practical application of NCOs to detect unmeasured confounding. To the authors' knowledge, only one previous study applied causal forests and NCOs for bias detection, but ours is the first systematic evaluation and validation of NCOs in causal ML.[29] Additionally, previous studies have tried to debias causal ML models by removing effects of unmeasured confounding for valid HTE estimation, but relied on other assumptions about transportability,[30] instrumental variables,[31, 32] or the data-generating process.[33] Other methodological studies have proposed approaches to predict interval bounds for ITE under violations of the no confounding assumption.[34-36] Our approach deviates from the literature by systematically characterizing how unmeasured confounding manifests across subgroups of a population in the presence and absence of true HTE.

Our study has several limitations. First, we applied a single set of input parameters for the prevalence of treatment and outcome. A lower treatment prevalence, resulting in greater treatment group imbalance, could theoretically alter the relative performance of the chosen ML models and the NCO. However, prior studies suggest that both causal forest and X-learner models perform well under treatment group imbalances[11, 37, 38], leading us to believe our results are robust to changing prevalence. Future work might focus on direct comparisons in the presence of unmeasured confounding and the utility of NCOs in these situations. Second, we did not assess the minimal event rate needed for an NCO to successfully detect unmeasured confounding. NCOs are most effective when the event is sufficiently common,[26] highlighting the need for researchers to carefully consider the variable choice when leveraging NCOs as a diagnostic tool. Third, we included only one unmeasured confounder in our simulations. In real-world observational data, confounding is often complex, involving multiple unmeasured covariates. However, we modeled this covariate with a relatively strong effect on both the treatment and outcome to approximate the influence of multiple unmeasured factors. Fourth, we simulated only linear relationships in the data generating models. Nonlinear relationships, better represent real-world data and could potentially improve the performance of causal forest models relative to the X-learner, as previous work has suggested.[38] Finally, we used LASSO and 10-fold cross-validation for estimating both the propensity score and outcome models in the X-learner. Alternative modeling approaches (i.e., changing or adding base learners) might better control for confounding.

    In conclusion, we demonstrate that conditional exchangeability violations pose a significant challenge in accurately estimating ITEs from routinely collected observational data using causal ML models. NCOs are a useful diagnostic tool for detecting local unmeasured confounding to assess potential bias in these analyses. Incorporating NCOs into standard diagnostic workflow in HTE estimation may help to increase the rigor of causal ML in health care research.

| Variable | Type | Prevalence/ Mean (SD) | Treatment Model Coefficient ($\beta_j$) | Outcome Model Coefficient ($\beta_k$) | NCO Model Coefficient ($\beta_{Nk}$) |
|---|---|---|---|---|---|
| **Table 1.** Simulation parameter settings in data-generating models ||||||
| **C1** | binary | 0.21 | -0.28 | 0.47 | 0.47 |
| **C2** | continuous | 77 (7.6) | 0.03 | 0.08 | 0.08 |
| **C3** | binary | 0.03 | 0.17 | 1.04 | 1.04* |
| **C4** | binary | 0.21 | 0 | -0.69 | -0.69 |
| **C5** | binary | 0.08 | 0 | 0.23 | 0.23* |
| **C6** | continuous | 8 (3) | 0.01 | 0 | 0 |
| **C7** | binary | 0.14 | 0.05 | 0 | 0 |
| **C8** | binary | 0.35 | 0.45 | 0 | 0 |
| **C9** | binary | 0.04 | 0.55 | 0 | 0 |
| **C10** | binary | 0.28 | -0.16 | 0 | 0 |
| **C11** | binary | 0.18 | 0.04 | 0.06 | 0.06 |
| **C12** | binary | 0.15 | 0.09 | 0.24 | 0.24 |
| **C13** | binary | 0.07 | 0.55 | -0.13 | -0.13 |
| **U** | binary | 0.10 | 1.50 | 1.50 | 1.50 |
| **Additional covariates for relaxed NCO simulation** ||||||
| **C14** | binary | 0.20 | 0.08 | 0 | 0.65 |
| **C15** | binary | 0.12 | 0.12 | 0 | -0.50 |
| **C16** | binary | 0.08 | -0.06 | 0 | 0.45 |

*Coefficients were set to 0 in the NCO-generating model in the scenario for simulating relaxed assumptions regarding NCO

| Simulation Setting | | | No Unmeasured Confounding | | Unmeasured Confounding | |
|---|---|---|---|---|---|---|
| | Measure | Oracle Model | Causal forest | X-learner | Causal forest | X-learner |
| **1. Primary Settings** | Q1 Mean Benefit | -0.158 | -0.119 | -0.150 | -0.049 | -0.079 |
| | Q2 Mean Benefit | -0.049 | -0.059 | -0.054 | -0.038 | -0.038 |
| | Q3 Mean Benefit | -0.037 | -0.048 | -0.039 | -0.033 | -0.026 |
| | Q4 Mean Benefit | -0.024 | -0.037 | -0.027 | -0.028 | -0.012 |
| | RMSE (95% CI) | NA | 0.037 (0.027, 0.054) | 0.030 (0.023, 0.039) | 0.064 (0.053, 0.076) | 0.055 (0.045, 0.066) |
| | ATE (95% CI) | -0.067 (-0.082, -0.053) | -0.066 (-0.080, -0.051) | -0.067 (-0.083, -0.052) | -0.037 (-0.052, -0.022) | -0.038 (-0.052, -0.024) |
| **2. Weaker Unmeasured Confounding** | Q1 Mean Benefit | -0.154 | -0.120 | -0.146 | -0.110 | -0.137 |
| | Q2 Mean Benefit | -0.050 | -0.061 | -0.058 | -0.059 | -0.056 |
| | Q3 Mean Benefit | -0.039 | -0.049 | -0.041 | -0.048 | -0.04 |
| | Q4 Mean Benefit | -0.026 | -0.038 | -0.026 | -0.039 | -0.027 |
| | RMSE (95% CI) | NA | 0.034 (0.025, 0.048) | 0.028 (0.021, 0.036) | 0.037 (0.025, 0.053) | 0.029 (0.022, 0.037) |
| | ATE (95% CI) | -0.067 (-0.082, -0.053) | -0.067 (-0.079, -0.053) | -0.068 (-0.082, -0.052) | -0.064 (-0.079, -0.050) | -0.064 (-0.079, -0.051) |
| **3. Sample size of 5,000** | Q1 Mean Benefit | -0.160 | -0.089 | -0.147 | -0.044 | -0.081 |
| | Q2 Mean Benefit | -0.053 | -0.067 | -0.062 | -0.037 | -0.042 |
| | Q3 Mean Benefit | -0.040 | -0.059 | -0.044 | -0.034 | -0.026 |
| | Q4 Mean Benefit | -0.023 | -0.051 | -0.020 | -0.030 | -0.006 |
| | RMSE (95% CI) | NA | 0.056 (0.034, 0.079) | 0.045 (0.030, 0.064) | 0.071 (0.046, 0.098) | 0.066 (0.042, 0.088) |
| | ATE (95% CI) | -0.069 (-0.094, -0.038) | -0.067 (-0.095, -0.036) | -0.068 (-0.096, -0.038) | -0.036 (-0.063, -0.007) | -0.038 (-0.066, -0.006) |

**Table 2.** Predicted treatment effect estimates and RMSE of predictions under a scenario of treatment benefit with true HTE across various simulation parameter settings

**Figure 1.** Subgroup conditional average treatment effect (CATE) estimates and 95% CIs on negative control outcome under a scenario of treatment benefit with true HTE across various simulation parameter settings

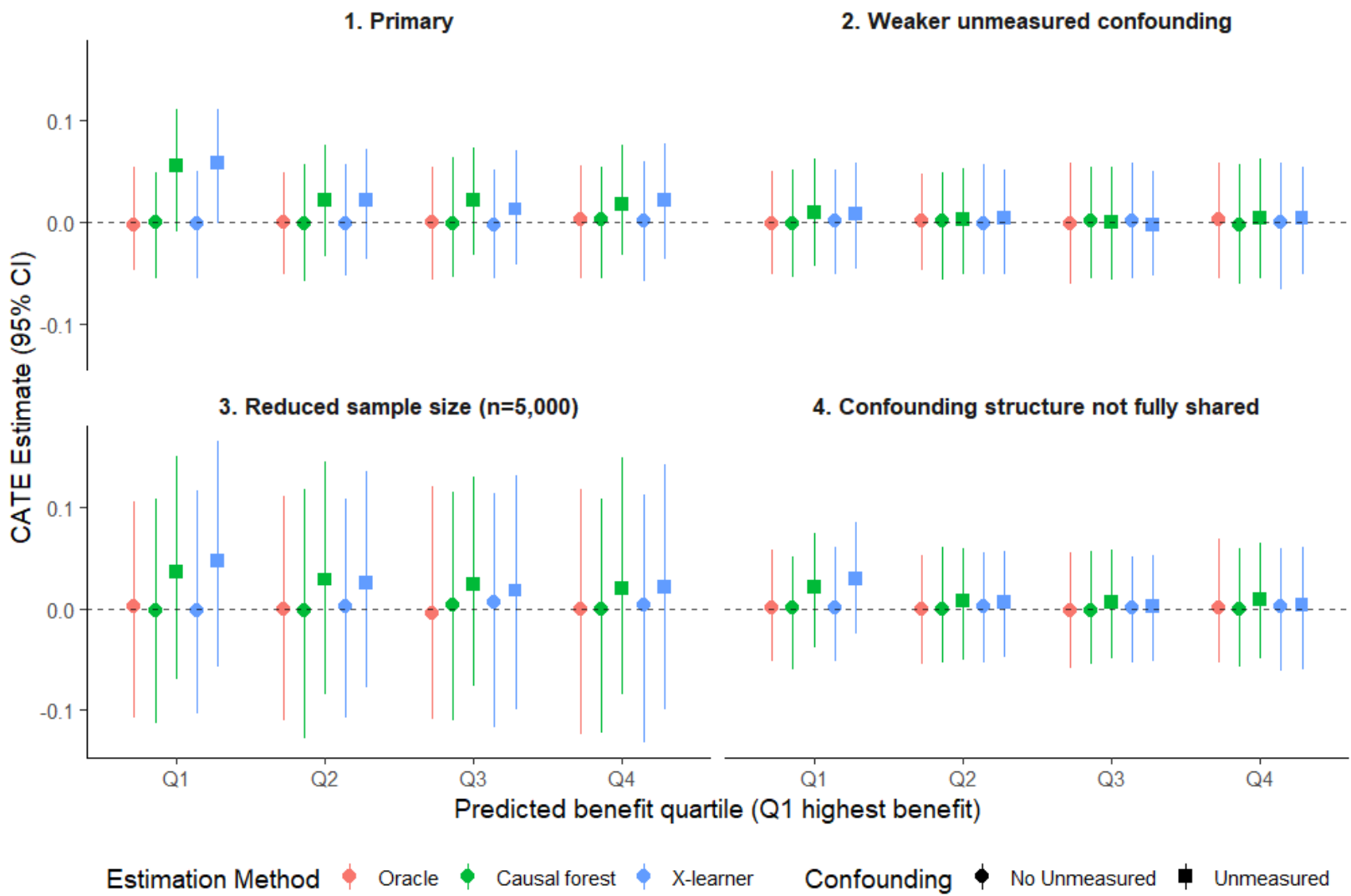

| Simulation Setting | | | No Unmeasured Confounding | | Unmeasured Confounding | |
|---|---|---|---|---|---|---|
| | Measure | Oracle Model | Causal forest | X-learner | Causal forest | X-learner |
| **1. Primary Settings** | Q1 Mean Benefit | -0.057 | -0.040 | -0.050 | -0.016 | -0.040 |
| | Q2 Mean Benefit | -0.040 | -0.037 | -0.038 | -0.010 | -0.022 |
| | Q3 Mean Benefit | -0.030 | -0.035 | -0.033 | -0.004 | -0.008 |
| | Q4 Mean Benefit | -0.017 | -0.032 | -0.022 | 0.012 | 0.047 |
| | RMSE (95% CI) | NA | 0.017 (0.010, 0.026) | 0.020 (0.011, 0.032) | 0.037 (0.026, 0.048) | 0.050 (0.035, 0.063) |
| | ATE (95% CI) | -0.036 (-0.051, -0.022) | -0.036 (-0.050, -0.023) | -0.036 (-0.051, -0.021) | -0.004 (-0.019, 0.011) | -0.006 (-0.021, 0.010) |
| **2. Weaker Unmeasured Confounding** | Q1 Mean Benefit | -0.058 | -0.041 | -0.051 | -0.037 | -0.047 |
| | Q2 Mean Benefit | -0.042 | -0.038 | -0.039 | -0.034 | -0.037 |
| | Q3 Mean Benefit | -0.032 | -0.036 | -0.035 | -0.032 | -0.032 |
| | Q4 Mean Benefit | -0.019 | -0.033 | -0.024 | -0.029 | -0.021 |
| | RMSE (95% CI) | NA | 0.016 (0.010, 0.024) | 0.019 (0.011, 0.029) | 0.017 (0.010, 0.026) | 0.019 (0.011, 0.030) |
| | ATE (95% CI) | -0.037 (-0.052, -0.025) | -0.037 (-0.050, -0.023) | -0.037 (-0.053, -0.023) | -0.033 (-0.049, -0.019) | -0.034 (-0.048, -0.021) |
| **3. Sample size of 5,000** | Q1 Mean Benefit | -0.069 | -0.041 | -0.063 | -0.015 | -0.045 |
| | Q2 Mean Benefit | -0.043 | -0.037 | -0.041 | -0.007 | -0.021 |
| | Q3 Mean Benefit | -0.031 | -0.034 | -0.031 | -0.001 | -0.005 |
| | Q4 Mean Benefit | -0.009 | -0.030 | -0.011 | 0.012 | 0.049 |
| | RMSE (95% CI) | | 0.029 (0.013, 0.050) | 0.035 (0.018, 0.057) | 0.046 (0.027, 0.069) | 0.055 (0.031, 0.083) |
| | ATE (95% CI) | -0.038 (-0.064, -0.008) | -0.036 (-0.062, -0.007) | -0.036 (-0.064, -0.006) | -0.002 (-0.029, 0.025) | -0.006 (-0.034, 0.023) |

**Table 3.** Predicted treatment effect estimates and RMSE of predictions under a scenario of treatment benefit with no HTE across various simulation parameter settings

**Figure 2.** Subgroup conditional average treatment effect (CATE) estimates and 95% CIs on negative control outcome under a scenario of treatment benefit with no HTE across various simulation parameter settings

## FUNDING

None.

## COMPETING INTERESTS

No conflicts of interest related to this work.